\documentclass[conference]{IEEEtran}
\IEEEoverridecommandlockouts
\usepackage[T1]{fontenc}
\usepackage{amsmath,amssymb,bm}
\usepackage{graphicx,booktabs,array}
\usepackage{siunitx,xcolor,cite,microtype}
\usepackage{algorithm,algorithmic,stfloats}

\usepackage{tikz}
\usetikzlibrary{svg.path}

\definecolor{linkpink}{RGB}{214,51,132}

\usepackage[
colorlinks=true,
linkcolor=black,
citecolor=black,
urlcolor=black
]{hyperref}
\graphicspath{{figures/}}
\definecolor{orcidlogocolor}{HTML}{A6CE39}
\tikzset{
	orcidlogo/.style={
		insert path={
			vertex [circle, fill=orcidlogocolor, minimum size=10pt, inner sep=0pt] at (0,0)
			node [white] at (0,0) {\sffamily\bfseries\tiny iD}
		}
	}
}

\newcommand{\orcidicon}[1]{%
	\href{https://orcid.org}{%
		\begin{tikzpicture}[baseline=-0.5ex]
			\draw[orcidlogocolor, fill=orcidlogocolor] (0,0) circle [radius=0.13];
			\node at (0,0) {\textcolor{white}{\sffamily\bfseries\tiny iD}};
		\end{tikzpicture}%
	}%
}

\title{\LARGE\bfseries \texttt{Qubit-to-Motion}: Traceable Action Provenance in Hybrid QRL for Robotic Bin-Packing Sequencing}
 \author{
 	\IEEEauthorblockN{Hassen Nigatu\textsuperscript{\orcidicon{0000-0002-4656-2725}1} and Howard Li\textsuperscript{\orcidicon{0000-0002-6359-0349}2}}
 	\IEEEauthorblockA{\textsuperscript{1}Zhejiang University Robotics Institute, Ningbo, China}
 	\IEEEauthorblockA{\textsuperscript{2}Department of Electrical and Computer Engineering, University of New Brunswick, Fredericton, NB, Canada}
 }
 
\hypersetup{pdftitle={Qubit-to-Motion: Traceable Action Provenance in Hybrid QRL for Robotic Bin-Packing Sequencing},pdfsubject={Revised manuscript},pdfkeywords={hybrid quantum reinforcement learning, action provenance, robotic placement}}
\begin{document}
\raggedbottom
\maketitle
\begin{abstract}
	Hybrid quantum reinforcement learning has been explored for robotic decision-making, but evaluations often emphasize task success, return, or convergence, leaving unresolved whether quantum circuit outputs influence constrained actions and remain traceable through robot execution. We present \texttt{Qubit-to-Motion}, an evidence-centered framework for support-constrained sequencing and simulated robotic placement. Four Pauli-$Z$ readouts provide the sole learned input to the actor and critic, while a support mask enforces admissible object selection. Each decision is recorded and linked via step indices and SHA-256 fingerprints to a sealed route, deterministic Cartesian and inverse-kinematics waypoints, prescribed joint motion, and MuJoCo contact replay. A frozen checkpoint is evaluated across five deterministic 30-object scene variants. All 150 placements satisfy support, containment, and settling criteria. A fixed-state zero-readout intervention changes $90.0\pm0.0\%$ of decisions, demonstrating dependence on the joint circuit representation. Mean macro travel is $34.537\pm0.206$~m, inference requires $8.60\pm1.27$~ms per route, and replay achieves $8.764\pm0.220~\mu$m mean final position error with 0.356~mm maximum interpenetration. These results establish reproducible decision-to-contact traceability. The implementation is available at \href{https://anonymous.4open.science/r/Qubit-to-Motion-Quantum-Reinforcement-Learning-F7DC}{\textcolor{linkpink}{the anonymous repository}}.
\end{abstract}


\section{Introduction}\label{sec:introduction}
\IEEEPARstart{S}{uccessful} robotic execution can conceal how learned computation shapes the decisions that produce it. In support-constrained packing, each object selection is made from a state-dependent admissible set and then propagated through route construction, robot motion, and contact execution. Two questions therefore remain: \emph{which learned representation influenced the selected action}, and \emph{can that action be identified unambiguously throughout the downstream execution record}? Existing robotic packing methods primarily evaluate geometric feasibility, stability, sequence quality, or task completion~\cite{wang2019stable,zhao2021online,huang2022planning,zhao2022feasible,zhang2025masking}, while hybrid quantum reinforcement-learning (QRL) studies commonly report return, success, or runtime~\cite{chen2020vqc,jerbi2021policies,skolik2022quantumagents,hohenfeld2024quantum}. These metrics are necessary, but they do not isolate decision-level dependence or establish continuity from learned representation to action and physical outcome.

Hybrid QRL makes the first question nontrivial because a classical encoder, a variational quantum circuit (VQC), a classical actor, and an invalid-action mask jointly determine the choice. The mask may force an action or alter the competition among actor logits~\cite{huang2022masking}; aggregate success therefore cannot show whether the circuit-derived readout affected a particular decision. A valid diagnostic must hold the logged observation, admissible set, actor, and tie-breaking rule fixed while changing only the readout vector. This tests readout sensitivity, not quantum advantage. The second question arises when the selected object is serialized into a route, expanded into waypoints, converted through inverse kinematics (IK) into joint motion, released into contact simulation, and recorded in logs, states, and visualizations. Persistent identifiers, route sealing, manifest references, and file-specific digests are needed to prevent this decision-to-contact link from being lost~\cite{pineau2021reproducibility,lebo2013provo,nist2015sha}.

\begin{figure}[!t]
	\centering
	\includegraphics[width=0.94\columnwidth]{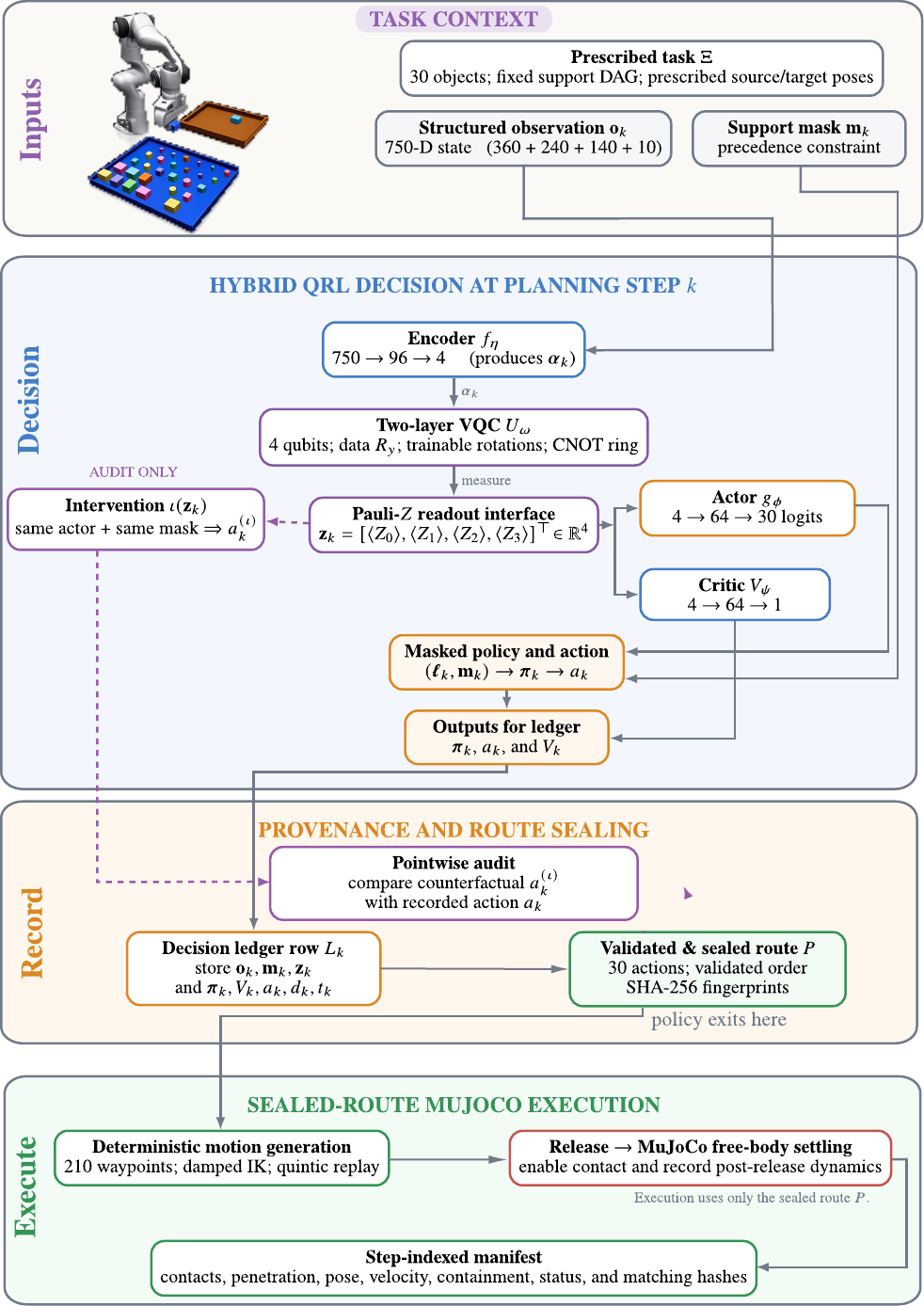}
	\caption{Decision-to-contact traceability in \texttt{Qubit-to-Motion}. Fixed-state substitutions probe the support-masked action at the original logged state. The validated route is then sealed, and common planning-step and object identifiers connect the action to waypoint generation, prescribed motion, free-body settling, and the resulting execution records.}
	\label{fig:overview}
\end{figure}

We address these coupled problems in the deliberately bounded setting of support-constrained sequencing for prescribed placements. Each object has a fixed source pose, target pose and yaw, dimensions, grasp offset, and set of support predecessors; the learned policy chooses only which currently admissible object to place next. We present \texttt{Qubit-to-Motion}, in which a trainable classical encoder drives a simulated two-layer, four-qubit VQC, and the four Pauli-$Z$ expectation values are the only features received by the actor and critic. A support mask enforces precedence, and behavior cloning followed by proximal policy optimization trains the policy~\cite{schulman2017ppo}. Fixed-state readout substitutions probe each logged action without changing the observation, mask, actor, or deterministic tie-breaking rule. After route validation and sealing, shared planning-step and object identifiers carry the action through seven-key waypoint generation, IK, prescribed Panda motion, MuJoCo free-body settling, placement records, saved states, and renderings~\cite{todorov2012mujoco}. Figure~\ref{fig:overview} summarizes the workflow.

The contributions of this work are threefold:
\begin{itemize}
	\item \textbf{Fixed-state diagnosis of a support-masked hybrid policy.} We intervene at the four-readout actor interface while preserving the original observation, mask, actor parameters, and tie-breaking rule. We quantify masked action changes and $L_1$ probability shifts for all decisions and for states with multiple admissible actions.
	\item \textbf{Decision-to-contact action provenance.} A decision ledger, route validation and sealing, persistent step and object identifiers, manifest references, and file-specific SHA-256 checks link policy inference to route entries, waypoints, placements, state histories, and visualizations, with forward and reverse trace queries.
	\item \textbf{End-to-end evaluation under controlled simulation.} For one checkpoint and five geometry-perturbed 30-object variants with a fixed support topology, we evaluate support validity, readout sensitivity, macro-travel, IK residuals, contact outcomes, and record continuity across 150 decisions, 1,050 waypoints, and 150 placements.
\end{itemize}

The remainder of this paper is organized as follows. Section~\ref{sec:related} reviews hybrid quantum policies, decision diagnostics and provenance, and packing sequence selection. Section~\ref{sec:methodology} presents the support-constrained MDP, actor--critic, training procedure, interventions, provenance records, motion pipeline, and computational analysis. Section~\ref{sec:experiments} describes the scenes, baselines, simulation setup, and aggregation protocol. Section~\ref{sec:results} reports the sequencing, intervention, kinematic, contact, and continuity results. Section~\ref{sec:traceability} presents trace queries and staged reproduction, and Section~\ref{sec:conclusion} concludes the paper.

\section{Related Work}\label{sec:related}
\subsection{Hybrid Quantum Policies and Robotic Applications}
Parameterized quantum circuits have been used as trainable value functions and policies~\cite{chen2020vqc,jerbi2021policies,skolik2022quantumagents}, including hybrid compression of high-dimensional observations~\cite{lockwood2021atari}. Robotic applications include navigation~\cite{hohenfeld2024quantum,dragan2025continuous}, autonomous driving~\cite{sinha2025navq}, and manipulator optimization~\cite{nigatu2025grover}; quantum learning and planning in robotics are also surveyed in~\cite{nigatu2026qmlrobotics}. Trainability, measurement, noise, and quantum--classical computation are established considerations in variational methods~\cite{preskill2018nisq,cerezo2021vqa}. We use a four-qubit analytic state-vector model and examine how its readouts affect constrained object selection.

\subsection{Decision Diagnostics and Provenance}
Explainable reinforcement learning and quantum-machine-learning methods attribute outputs to inputs, events, or circuit components~\cite{beechey2023sverl,brindise2024diagnostics,heese2025xqml}. Invalid-action masking defines the admissible support of a policy distribution~\cite{huang2022masking}. We therefore preserve the logged mask when comparing the original and intervened readouts. Reproducibility studies address experimental seeds, statistical reporting, and computational records~\cite{henderson2018matters,agarwal2021precipice,pineau2021reproducibility}, while PROV-O represents provenance through entities and their derivation relations~\cite{lebo2013provo}. Our framework joins a local policy diagnostic to explicit action identities across motion and contact records.

\subsection{Packing and Placement Sequencing}
Classical three-dimensional bin packing considers object positions and orientations~\cite{martello2000three,crainic2008extreme}. Robotic packing further incorporates stability and execution~\cite{wang2022dense}, and demonstrations can guide the selection of packing sequences~\cite{santos2024sequence}. We focus on the support-valid topological order of prescribed placements. Fixed target geometry and a shared motion template let us compare sequencing decisions and follow each selected object through a consistent execution process.

\section{Methodology} \label{sec:methodology}
\subsection{Finite-Horizon MDP and Support Constraints}
We represent a fixed scene by $\Xi$ and its support relations by the directed acyclic graph $G=(\mathcal V,\mathcal E)$, where $\mathcal V=\{0,\ldots,N-1\}$ and $N=30$. Before decision $k$, the committed prefix is $P_k=(a_0,\ldots,a_{k-1})$ and its selected-object set is $C_k$. Our state is $s_k=(P_k,C_k,\Xi)$. With $\operatorname{pred}(i)=\{j:(j,i)\in\mathcal E\}$, we define
\begin{align}
\mathcal A(s_k)&=\{i\in\mathcal V\setminus C_k:\operatorname{pred}(i)\subseteq C_k\},\label{eq:admissible}\\
m_{k,i}&=\mathbf1[i\notin C_k\wedge\operatorname{pred}(i)\subseteq C_k],\label{eq:mask}\\
T(s_k,i)&=(P_k\Vert i,C_k\cup\{i\},\Xi).\label{eq:transition}
\end{align}
We terminate after all $N$ objects have been selected. The prefix length supplies the decision index, giving the finite-horizon MDP $\mathcal M_\Xi=(\mathcal S,\mathcal V,T,r,\gamma,N)$ with the state-dependent admissible sets in~\eqref{eq:admissible}.

We measure order-level geometry using the source center $\mathbf s_i$ and target center $\mathbf x_i$ of object $i$. Let $\mathbf c_0$ be the home tool position and $\mathbf c_k=\mathbf x_{a_{k-1}}$ for $k>0$. Our incremental macro-travel cost and complete-route surrogate are
\begin{align}
d_k(i)&=\|\mathbf c_k-\mathbf s_i\|_2+\|\mathbf s_i-\mathbf x_i\|_2,\nonumber\\
J(P)&=\sum_{k=0}^{N-1}d_k(a_k).\label{eq:travel}
\end{align}
Every complete route includes the same source-to-target distance sum, $\sum_{i\in\mathcal V}\|\mathbf s_i-\mathbf x_i\|_2$. Consequently, route-to-route differences in $J$ arise from the transitions from the preceding target, or the initial home position, to the next source. We use $J$ to compare object orders; the seven-key motion template in Section~\ref{sec:motion} specifies the executed trajectory.

Our shaped reward, with distances expressed in metres, is
\begin{equation}
r_k=1-0.65d_k(a_k)+0.12\,\mathbf1[a_k=a_k^{\mathrm g}]
       +5\,\mathbf1[k=N-1],\label{eq:reward}
\end{equation}
where $a_k^{\mathrm g}$ is the support-admissible action minimizing $d_k(i)$. We train for discounted return, whose value satisfies
\begin{align}
V^\pi(s_k)&=\sum_{i\in\mathcal A(s_k)}\pi(i\mid s_k)
 \bigl[r(s_k,i)+\gamma V^\pi(T(s_k,i))\bigr],\label{eq:bellman}\\
V^\pi(s_N)&=0.\nonumber
\end{align}
The per-step constant and terminal bonus contribute the same discounted amount to every complete valid route. Travel cost and greedy-action agreement determine the route-dependent training signal. Our reported $J$ is an undiscounted geometric summary, whereas the training objective weights the shaped rewards by $\gamma^k$.

\subsection{Hybrid Quantum Actor--Critic}
We construct $\mathbf o_k\in[-1,1]^{750}$ by concatenating $30\times12$ object features, $30\times8$ placement/support features, a $14\times10$ committed-height map, and ten global features. Our trainable 750--96--4 encoder $f_\eta$ produces $\mathbf e_k\in[-1,1]^4$, from which we form $\boldsymbol\alpha_k=(\pi/2)\mathbf e_k$.

We evaluate a two-layer, four-qubit variational quantum circuit (VQC) using analytic state-vector simulation. The circuit uses data $R_y$ rotations, trainable $R_xR_yR_z$ rotations, and the CNOT ring $(0,1)(1,2)(2,3)(3,0)$. For circuit parameters $\omega$ and the resulting state $\rho_k$, our readout vector is
\begin{equation}
\mathbf z_k=[z_{k,0},\ldots,z_{k,3}]^\top,\qquad
z_{k,q}=\operatorname{Tr}(\rho_k Z_q).\label{eq:readout}
\end{equation}
Each $Z_q$ denotes the Pauli-$Z$ observable on qubit $q$, and each $z_{k,q}$ is its real-valued expectation. We feed $\mathbf z_k$ to both the 4--64--30 actor $g_\phi$ and the 4--64--1 critic $v_\psi$. Thus, the circuit readouts are the sole learned features received by the two heads; the support mask separately encodes the discrete precedence constraints.

\begin{figure*}[!t]
\centering
\includegraphics[width=0.98\textwidth]{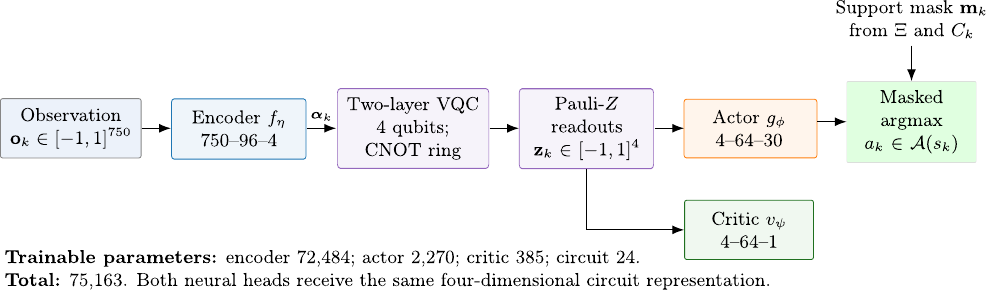}
\caption{Our hybrid actor--critic and parameter counts. We pass the same four circuit readouts to both neural heads and apply the support mask to the actor's logits. The encoder, circuit, actor, and critic contain 75,163 trainable parameters in total.}
\label{fig:architecture}
\end{figure*}

Writing $\boldsymbol\ell_k=g_\phi(\mathbf z_k)$ and $\theta=(\eta,\omega,\phi,\psi)$, we form
\begin{equation}
\pi_\theta(i\mid\mathbf o_k,\mathbf m_k)
=\frac{m_{k,i}\exp(\ell_{k,i})}{\sum_{j=0}^{N-1}m_{k,j}\exp(\ell_{k,j})}.\label{eq:policy}
\end{equation}
We implement masking with a large negative sentinel for invalid logits. During evaluation, we select
\begin{equation}
a_k=\min\operatorname*{arg\,max}_{i\in\mathcal A(s_k)}\ell_{k,i},\label{eq:action}
\end{equation}
so an exact tie is resolved by the smallest admissible index. Our four-qubit bottleneck provides four logged features per decision, while two circuit layers give 24 trainable rotation parameters. Figure~\ref{fig:architecture} details the associated neural-network counts.

\subsection{Training}
We use 540 support-valid demonstration decisions from 18 generated scenes and initialize our hybrid policy by BC against the greedy action labels $a_k^{\mathrm g}$. We then apply two PPO updates, each containing 120 planning decisions and three optimization passes. Table~\ref{tab:training} collects our training settings.

\begin{table}[!t]
\centering
\caption{Our training configuration.}
\label{tab:training}
\footnotesize
\setlength{\tabcolsep}{5pt}
\begin{tabular}{@{}lr@{}}
\toprule
Setting & Value\\
\midrule
Training seed & 2025\\
Demonstration scenes / decisions & 18 / 540\\
BC epochs / learning rate & 15 / $2\times10^{-3}$\\
PPO updates / decisions per update & 2 / 120\\
Optimization passes per PPO update & 3\\
Configured PPO batch size & 128\\
PPO learning rate & $3\times10^{-4}$\\
Discount $\gamma$ / GAE parameter $\lambda$ & 0.985 / 0.95\\
PPO clip ratio & 0.2\\
Entropy / value coefficients & 0.01 / 0.5\\
Gradient-norm limit & 0.7\\
\bottomrule
\end{tabular}
\end{table}

We minimize masked cross-entropy during BC. For PPO, we evaluate the critic as $\widehat V_k=v_\psi(\mathbf z_k)$ and set the terminal bootstrap value to $\widehat V_N=0$. Our temporal-difference residuals and generalized advantage estimates (GAE)~\cite{schulman2016gae} are
\begin{align}
\epsilon_k&=r_k+\gamma\widehat V_{k+1}-\widehat V_k,\label{eq:td}\\
\widehat A_k&=\sum_{j=0}^{N-k-1}(\gamma\lambda)^j\epsilon_{k+j}.\label{eq:gae}
\end{align}
The PPO probability ratio compares the updated and rollout policies on the same recorded observation, mask, and action. We use the clipped policy objective with $\widehat A_k$, fit the critic to the corresponding return targets, and regularize the masked policy with entropy. The greedy demonstrations and agreement term in~\eqref{eq:reward} provide a common reference throughout training.

\subsection{Fixed-State Readout Interventions}\label{sec:intervention}
We evaluate representation sensitivity at the actor input. Let $\bar\pi_\phi(\cdot\mid\mathbf z,\mathbf m)$ denote the distribution obtained by passing $\mathbf z$ through the actor and applying mask $\mathbf m$. For a readout transformation $\iota:\mathbb R^4\to\mathbb R^4$, we compute
\begin{align}
a_k^{(\iota)}&=\min\operatorname*{arg\,max}_{i\in\mathcal A(s_k)}
 \bar\pi_\phi(i\mid\iota(\mathbf z_k),\mathbf m_k),\label{eq:intervention}\\
\Delta_k^{(\iota)}&=\mathbf1[a_k^{(\iota)}\ne a_k].\nonumber
\end{align}
We preserve the original observation, mask, actor parameters, and tie-breaking rule at every comparison. For each evaluated route, we summarize
\begin{align}
D_{\mathrm{all}}^{(\iota)}&=\frac{1}{N}\sum_{k=0}^{N-1}\Delta_k^{(\iota)},\label{eq:sensitivity}\\
D_{\mathrm{choice}}^{(\iota)}&=
\frac{\sum_{k\in\mathcal K}\Delta_k^{(\iota)}}{|\mathcal K|},\qquad
\mathcal K=\{k:|\mathcal A(s_k)|>1\}.\nonumber
\end{align}
All evaluated routes have a nonempty $\mathcal K$. A one-action mask makes the selected action invariant to the readout substitution.

We replace the entire readout vector with zero, set each component $z_{k,q}$ individually to zero, reverse the component order, and negate the vector. The zero-vector input yields a fixed actor-logit vector, which each logged mask restricts to its admissible entries. We report action-change rates together with the mean $L_1$ probability shifts. At an individual state, this shift is
\begin{equation}
\Delta_{1,k}^{(\iota)}=
\bigl\|\bar\pi_\phi(\cdot\mid\iota(\mathbf z_k),\mathbf m_k)
-\bar\pi_\phi(\cdot\mid\mathbf z_k,\mathbf m_k)\bigr\|_1.\label{eq:l1}
\end{equation}
Our pointwise statistics use the original logged states. We also generate a separately identified recursive zero-readout visualization in which altered selections update subsequent states.

\subsection{Decision Ledger and Route Identity}\label{sec:identity}
We record each planning decision as
\begin{equation}
L_k=(k,\mathbf o_k,\mathbf m_k,\mathbf e_k,\mathbf z_k,
\boldsymbol\pi_k,\widehat V_k,a_k,d_k(a_k),t_k),\label{eq:ledger}
\end{equation}
where $t_k$ is the measured inference duration. The stored encoder output determines the circuit angles through $\boldsymbol\alpha_k=(\pi/2)\mathbf e_k$. After planning, we validate route length, object uniqueness, support precedence, and ledger indices, then seal the route for execution.

We assign a SHA-256 digest to each file and compare its recomputed digest with the recorded reference for that same file~\cite{nist2015sha}. Manifest references express the dependencies between different files. We match the planning-step and object identifiers across the ledger, route, waypoint records, and placement records. In symbolic form, our identity check at placement index $k$ is
\begin{equation}
a_k^{\mathrm{ledger}}=a_k^{\mathrm{route}}
=a_k^{\mathrm{motion}}=a_k^{\mathrm{placement}}.\label{eq:identity}
\end{equation}
We complete the ledger and route validation before execution. Our motion and simulator stages then consume the sealed object order, while the manifest connects their outputs to the corresponding input digests. Together, these checks associate the simulated placement with the action selected at the same planning index.

\subsection{Deterministic Motion and MuJoCo Settling}\label{sec:motion}
We expand each selected object into seven Cartesian keys: home, pregrasp, grasp, lift, preplace, place, and retreat. We solve damped position IK for the seven Panda joints and enforce joint limits. Between successive joint configurations, we use quintic time scaling
\begin{equation}
\sigma(\tau)=10\tau^3-15\tau^4+6\tau^5,\qquad 0\leq\tau\leq1,\label{eq:quintic}
\end{equation}
with normalized segment time $\tau$. Each route produces 210 waypoint records.

We prescribe Panda joint motion kinematically and attach the selected object deterministically during transport. We disable robot collision geometries and selected-object contacts during that phase. We detach the object 1.5~mm above its prescribed target pose, restore its free-body contact state, and simulate contact and settling in MuJoCo 3.11.0. These stages retain the object identifier from the sealed route.

\begin{algorithm}[!t]
\caption{Our Sealed Sequencing and Placement Procedure}
\label{alg:pipeline}
\footnotesize
\begin{algorithmic}[1]
\REQUIRE Scene $\Xi$; trained parameters $\theta$; MuJoCo 3.11.0
\STATE Initialize $P\leftarrow()$, $C\leftarrow\emptyset$, and ledger $\mathcal L$
\FOR{$k=0$ to $N-1$}
 \STATE Construct $\mathbf o_k$ and $\mathbf m_k$; require $\mathcal A(s_k)\ne\emptyset$
 \STATE Evaluate encoder, circuit, actor, critic, and masked action $a_k$
 \STATE Record $L_k$, append $a_k$ to $P$, and update $C$
\ENDFOR
\STATE Validate route and ledger; record their digests and seal $P$
\FOR{$k=0$ to $N-1$}
 \STATE Generate seven Cartesian keys for $P[k]$ and solve position IK
 \STATE Execute the quintic segments with attachment during transport
 \STATE Detach $P[k]$ 1.5~mm above its target and enable free-body contact
 \STATE Simulate until 60 consecutive quiet samples or 500 steps
 \STATE Record contacts, pose, velocities, support, containment, and status
 \STATE Associate output records with $k$, $P[k]$, and input digests
\ENDFOR
\RETURN Waypoint records, placement records, state history, and manifest
\end{algorithmic}
\end{algorithm}

We terminate the settling interval after 60 consecutive quiet samples or a maximum of 500 simulation steps. A quiet sample requires linear speed below 5~mm/s and angular speed below 0.05~rad/s. We assess each placement using bin containment, contact with the prescribed support, position error below 5~mm, and final linear speed below 8~mm/s. Thus, the quiet-sample rule specifies the settling duration, while the placement gates specify the recorded outcome. Algorithm~\ref{alg:pipeline} summarizes the processing order.

\subsection{Computational Cost}
For a constructed observation, we express the forward-decision cost in terms of observation dimension $D$, encoder width $H$, qubit count $Q$, circuit depth $L$, head width $H_h$, and mean predecessor count $\bar s=|\mathcal E|/N$. With predecessor lists and direct selected-set membership checks, we obtain
\begin{align}
T_{\mathrm{dec}}=\mathcal O\bigl(&DH+HQ+LQ2^Q\nonumber\\
 &+QH_h+NH_h+N+|\mathcal E|\bigr),\label{eq:time}\\
S_{\mathrm{dec}}=\mathcal O\bigl(&P_{\mathrm{par}}+D+N+|\mathcal E|+2^Q\bigr),\label{eq:space}
\end{align}
where $P_{\mathrm{par}}$ is the stored parameter count. We include the $N$-object scan and predecessor checks in the mask cost. The state-vector term follows from applying $\mathcal O(LQ)$ one- and two-qubit gates to $2^Q$ amplitudes. A route contains $N$ forward decisions. Retaining all observation, mask, distribution, and readout entries adds $\mathcal O(N(D+N+Q))$ ledger storage, separate from the per-decision working memory and simulator-state histories.

\section{Experimental Configuration} \label{sec:experiments}
\subsection{Scenes and Baselines}
We evaluate one checkpoint trained with seed 2025 on scene seeds 41, 73, 109, 151, and 197. Each scene has 15 floor objects, indexed 0--14, and 15 upper objects; object $15+i$ depends on floor object $i$. We permute the source order and perturb the geometry using source-coordinate jitter bounded by $\pm3$~mm, target-center jitter bounded by $\pm2$~mm, half-size changes of up to 0.8~mm in-plane and 0.7~mm vertically, and yaw within $\pm0.12$~rad. The checkpoint and support topology remain fixed across these five seeded variants.

We compare our hybrid routes with three sequencing baselines under the same scene geometry. Our nearest-greedy baseline selects the admissible object minimizing $d_k(i)$. Our canonical baseline selects the lowest-index admissible object. For the random-valid baseline, we sample uniformly from the current admissible set, commit the sampled action, and recompute the set. We generate 200 routes per scene using an independent pseudorandom stream and summarize them by their scene-level median. This sequential sampling procedure defines the random-route distribution used in our comparison.

\subsection{Simulation and Recording}
We use MuJoCo 3.11.0 with CPython 3.13. Our simulator uses the implicit-fast integrator, a 2~ms time step, an elliptic friction cone, 80 solver iterations, and 30 line-search iterations. Each box has mass 85~g and high-friction contacts. With our time step, the 60-sample quiet window spans 0.120~s and the 500-step settling limit spans 1.000~s.

For every placement, we record contact samples, penetration derived from negative contact distance, peak normal contact force, settling time, final linear and angular speed, position and yaw error, support identity, containment, and gate status. We save simulator-state histories and render them through EGL at $640\times360$ pixels. We apply Lanczos resampling to produce $1920\times1080$ H.264 video at 30 frames/s, and associate each video with its state-history and model references.

\subsection{Evaluation and Aggregation}
We evaluate route validity, fixed-state decision sensitivity, macro travel, IK key-position residuals, and placement outcomes. Our continuous summaries use the mean and scene-level standard deviation across the five evaluated variants. Position-error summaries first aggregate the 30 placements within each scene; waypoint summaries refer to the 210 Cartesian IK keys per scene. For the random-valid baseline, the across-scene summary aggregates the five scene-level medians. Decision-change rates use all 150 recorded decisions or the 145 observed non-forced decisions, as appropriate.

We check route cardinality, uniqueness, precedence, and ledger indices before sealing. We check waypoint cardinality and IK tolerances before replay, and assign placement status from the contact and state records. Each execution stage uses the sealed route and retains its step and object identifiers. A failed validation stops the affected stage and records its status. We assess video encoding and sampled-frame content separately from the placement gates.

\section{Results} \label{sec:results}
\subsection{Support-Valid Sequencing and Macro Travel}
We obtain five complete routes, each containing all 30 objects exactly once and satisfying the support graph. Table~\ref{tab:results} reports the scene-level route and placement values. Our hybrid macro-travel surrogate is $34.537\pm0.206$~m. The corresponding summaries are $34.437\pm0.119$~m for nearest greedy, $34.523\pm0.186$~m for canonical order, and $34.661\pm0.140$~m for the random-valid medians. The hybrid mean is 0.100~m above greedy, 0.014~m above canonical, and 0.124~m below the random-valid median summary.

\begin{table*}[!t]
\centering
\caption{Our scene-level sequence and placement results. $J_{\mathrm H}$ and $J_{\mathrm G}$ denote hybrid and nearest-greedy macro travel. $D_{\mathrm{all}}$ and $D_{\mathrm{choice}}$ refer to the fixed-state all-zero intervention.}
\label{tab:results}
\small
\setlength{\tabcolsep}{7pt}
\begin{tabular}{@{}rrrrrrrr@{}}
\toprule
Scene seed & $J_{\mathrm H}$ (m) & $J_{\mathrm G}$ (m) & $D_{\mathrm{all}}$ (\%) & $D_{\mathrm{choice}}$ (\%) & Position mean ($\mu$m) & Position max. ($\mu$m) & Passed / total\\
\midrule
41  & 34.499 & 34.348 & 90.0 & 93.1 & 8.883 & 16.170 & 30 / 30\\
73  & 34.804 & 34.536 & 90.0 & 93.1 & 8.544 & 16.158 & 30 / 30\\
109 & 34.678 & 34.572 & 90.0 & 93.1 & 9.050 & 16.174 & 30 / 30\\
151 & 34.283 & 34.437 & 90.0 & 93.1 & 8.544 & 16.156 & 30 / 30\\
197 & 34.421 & 34.293 & 90.0 & 93.1 & 8.796 & 16.165 & 30 / 30\\
\bottomrule
\end{tabular}
\end{table*}

In scene 151, our hybrid route has 0.154~m lower macro travel than nearest greedy. In the other four scenes, its macro travel is higher by 0.106--0.268~m. These comparisons describe the orders produced by our greedy-initialized, reward-shaped policy under the fixed scene geometry. Figure~\ref{fig:results}(a) visualizes the hybrid and greedy values in Table~\ref{tab:results}.

We can describe the support-mask evolution exactly for our paired, two-layer graph. If $f_k$ floor objects and $u_k$ upper objects have been committed, then $15-f_k$ floor objects and $f_k-u_k$ upper objects are admissible. Hence,
\begin{equation}
|\mathcal A(s_k)|=(15-f_k)+(f_k-u_k)=15-u_k.\label{eq:cardinality}
\end{equation}
Selecting a floor object preserves this cardinality by making its upper object available; selecting an upper object reduces it by one. In each of our five evaluated routes, we observe 29 non-forced decisions and one forced final decision. We interpret the selected-action probability relative to this changing admissible set.

\subsection{Fixed-State Decision Dependence}
\begin{figure*}[!t]
\centering
\begin{minipage}[t]{0.475\textwidth}
\centering
\includegraphics[width=\linewidth]{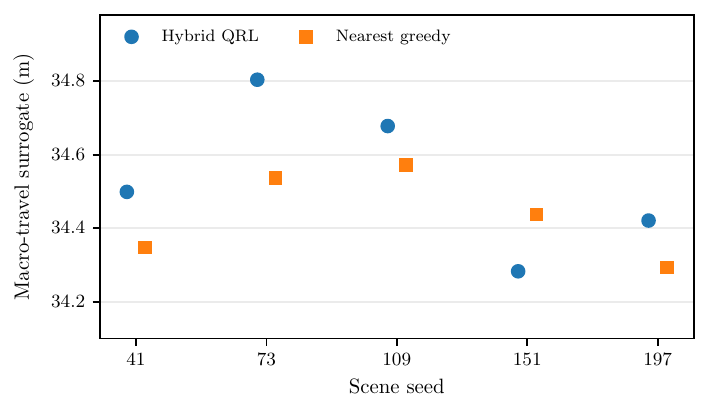}\\[-1pt]
{\footnotesize (a) }
\end{minipage}\hfill
\begin{minipage}[t]{0.505\textwidth}
\centering
\includegraphics[width=\linewidth]{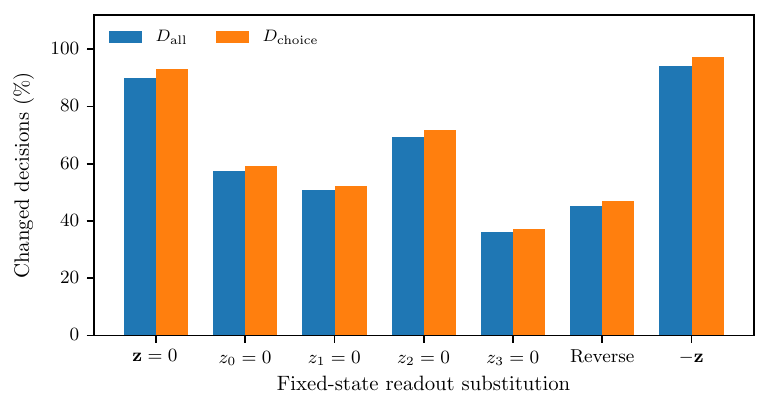}\\[-1pt]
{\footnotesize (b)}
\end{minipage}
\caption{Our tabulated sequencing and intervention results. Panel (a) uses the scene-level values in Table~\ref{tab:results}; panel (b) uses the percentages in Table~\ref{tab:interventions}. The intervention comparisons retain each original state and support mask.}
\label{fig:results}
\end{figure*}

Our all-zero intervention changes 27 of 30 decisions in every scene. Across the five routes, this gives $135/150=90.0\%$ of all decisions and $135/145\approx93.1\%$ of non-forced decisions. Table~\ref{tab:interventions} and Fig.~\ref{fig:results}(b) summarize the expanded interventions.

\begin{table}[!t]
\centering
\caption{Our fixed-state readout interventions. We use 150 decisions for $D_{\mathrm{all}}$ and 145 non-forced decisions for $D_{\mathrm{choice}}$. Each $z_q$ denotes a readout component.}
\label{tab:interventions}
\footnotesize
\setlength{\tabcolsep}{4pt}
\begin{tabular}{@{}lrrr@{}}
\toprule
Readout substitution & $D_{\mathrm{all}}$ (\%) & $D_{\mathrm{choice}}$ (\%) & Mean $L_1$\\
\midrule
$\mathbf z\leftarrow\mathbf0$ & 90.0 & 93.1 & 0.102\\
$z_0\leftarrow0$ & 57.3 & 59.3 & 0.136\\
$z_1\leftarrow0$ & 50.7 & 52.4 & 0.109\\
$z_2\leftarrow0$ & 69.3 & 71.7 & 0.135\\
$z_3\leftarrow0$ & 36.0 & 37.2 & 0.081\\
Reverse component order & 45.3 & 46.9 & 0.070\\
$\mathbf z\leftarrow-\mathbf z$ & 94.0 & 97.2 & 0.198\\
\bottomrule
\end{tabular}
\end{table}

Negating the vector changes 94.0\% of all decisions and gives the largest mean $L_1$ shift, 0.198, among our tested interventions. Individual-component zeroing changes 36.0--69.3\% of decisions. The action-change rate measures a crossing of the masked argmax boundary, while the $L_1$ shift measures redistribution of probability. Their relationship depends on the action margins and the actor's joint response to the four readouts. We therefore interpret these results as state-specific sensitivity of our implemented actor to its circuit-derived representation.

\subsection{Kinematics and Contact Outcomes}
We generate 210 key waypoints per route and 1,050 across the five scenes. Our mean end-effector key-position IK residual is $0.166\pm0.002$~mm, and the largest residual is 0.300~mm. Analytic state-vector policy inference takes $8.60\pm1.27$~ms per complete 30-decision route.

All 150 simulated placements pass the prescribed-support, containment, position, and final-speed gates. Our mean final object-position error is $8.764\pm0.220~\mu$m, and the largest recorded error is $16.174~\mu$m. The maximum yaw error is $0.000232^\circ$. These object-state measurements follow detachment 1.5~mm above the prescribed target and subsequent free-body settling. They quantify a different stage of the pipeline from the IK key-position residuals.

\begin{figure*}[!t]
\centering
\includegraphics[width=0.82\textwidth]{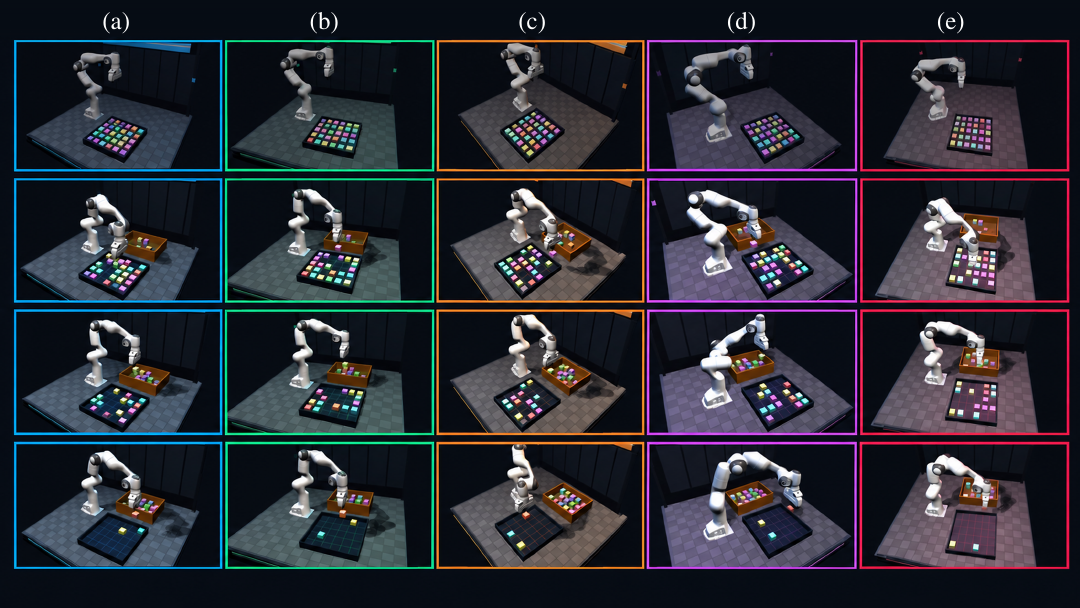}
\caption{We visualize the five scene configurations in columns (a)--(e) at matched simulation times. From top to bottom, the rows correspond to $t=1.0$, 7.5, 14.0, and 23.5~s. The views show our prescribed robot motion and the evolving source and target object configurations.}
\label{fig:workspace_final_states}
\end{figure*}

\begin{figure*}[!t]
\centering
\includegraphics[width=0.90\textwidth]{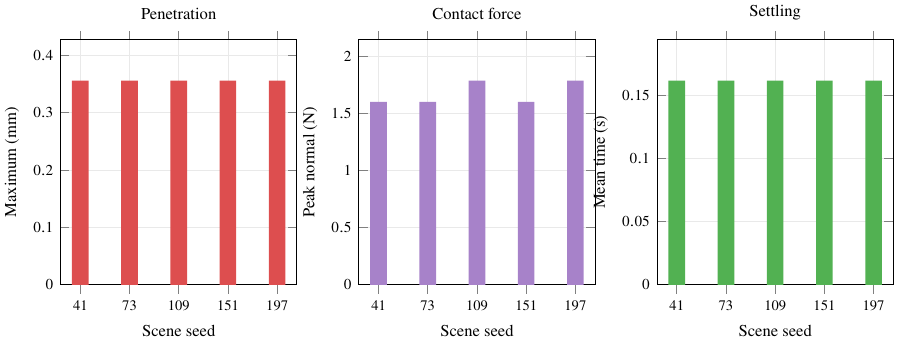}
\caption{Our per-scene contact measurements: maximum object--support penetration, peak normal contact force, and mean settling time. We obtain these quantities from the free-body contact interval after detachment.}
\label{fig:contact}
\end{figure*}

We record a maximum object--support penetration of 0.356~mm, a peak normal contact force of 1.79~N, and a mean settling time of 0.162~s. The scene-level standard deviation of settling time rounds to 0.000~s at the reported precision. Figure~\ref{fig:contact} shows the scene-level contact summaries. We illustrate the simulated manipulation progression in Fig.~\ref{fig:workspace_final_states}.

\subsection{Action-to-Contact Continuity}
We preserve each route's object order through its 210 waypoint records and 30 placement records. The manifest associates every placement with the same planning-step and object identifiers as the decision ledger. File-specific digest checks connect the checkpoint, scene, route, ledger, trajectory, state history, and rendering metadata through their recorded references.

We render five original-route videos and one recursive zero-readout video, and assemble a side-by-side comparison and a five-scene montage. All eight videos use H.264 High Profile at $1920\times1080$ pixels and 30 frames/s, and satisfy our sampled-frame nonblank check. Direct renderings reference their saved simulator states and model; composite videos reference their source videos. We use the contact and state records to determine placement outcomes and the videos to visualize the corresponding simulations.

\section{Traceability and Reproducible Execution} \label{sec:traceability}
\subsection{Forward and Reverse Trace Queries}
We support inspection in both directions. Starting from $L_k$, we follow the selected object to the same route index, its seven waypoint records, the detachment and contact records, and the saved state history. The video metadata identify the state history used for rendering. Starting from a placement record, we recover the originating action, support mask, actor distribution, and four circuit readouts. This bidirectional organization gives each reported placement a defined computational antecedent.

Our stage-specific checks localize processing failures. We validate the scene and checkpoint references at planning, route cardinality and precedence at sealing, and waypoint counts and IK tolerances before replay. We record simulator outcomes at placement and encoding outcomes at rendering. Each completed stage retains its inputs, outputs, and status, so we can inspect a later-stage failure alongside the records already produced.

\subsection{Staged Reproduction}
We organize reproduction into environment, policy, execution, and validation stages. We pin CPython and MuJoCo versions, compile a smoke-test model, exercise contact dynamics, and check EGL rendering. We regenerate the demonstration data, training records, checkpoint, scene specifications, routes, ledgers, and interventions in the policy stage. For each scene, we then construct its model, generate the 210-waypoint trajectory, execute 30 placements, and save the placement and state-history records.

We maintain the simulator distribution, Panda MJCF model and meshes, source code and dependency sources, checkpoint, scene specifications, decision records, waypoint records, placement logs, compressed state histories, and manifests. Our validation stage recomputes file digests, checks cardinalities and outcomes, and inspects video properties. The resulting record structure supports repeatable examination of the same checkpoint, five scene variants, prescribed transport procedure, and free-body settling configuration.

\section{Conclusion}\label{sec:conclusion}
We introduced \texttt{Qubit-to-Motion} to connect readout-dependent hybrid-QRL decisions with support-constrained simulated placement. Our framework combines a four-readout actor--critic, fixed-state interventions, a decision ledger, sealed object orders, deterministic Panda motion, and MuJoCo free-body settling. Across five 30-object scenes, we connect 150 decisions to 1,050 waypoints and 150 placements that satisfy our recorded outcome criteria. The all-zero intervention changes 135 of 150 actions, demonstrating decision dependence on the circuit-derived representation for the evaluated checkpoint. Through shared step and object identifiers and file-specific digest checks, we retain the connection from each selected action to its trajectory, contact record, and visualization. Our results establish an integrated procedure for examining hybrid-policy decisions and tracing their execution in simulation.

\end{document}